\documentclass[letterpaper]{article} 
\usepackage{aaai25}  
\usepackage{times}  
\usepackage{helvet}  
\usepackage{courier}  
\usepackage[hyphens]{url}  
\usepackage{graphicx} 
\usepackage{natbib}  
\usepackage{caption} 
\usepackage{amsmath}
\usepackage{graphicx}
\usepackage{subcaption}
\usepackage{booktabs}
\usepackage{amssymb}

\usepackage{algorithm}
\usepackage{algorithmic}

\usepackage{newfloat}
\usepackage{listings}
\DeclareCaptionStyle{ruled}{labelfont=normalfont,labelsep=colon,strut=off} 
\floatstyle{ruled}
\newfloat{listing}{tb}{lst}{}
\floatname{listing}{Listing}
\title{GCAD: Anomaly Detection in Multivariate Time Series from the \\Perspective of Granger Causality}
\author {
    Zehao Liu\textsuperscript{\rm 1},
    Mengzhou Gao\textsuperscript{\rm 1,\rm 2},
    Pengfei Jiao\textsuperscript{\rm 1,\rm 2,\rm 3}\thanks{Corresponding author}}
\affiliations {
    \textsuperscript{\rm 1}Zhuoyue Honors College, Hangzhou Dianzi University, China\\
    \textsuperscript{\rm 2}School of Cyberspace, Hangzhou Dianzi University, China\\
    \textsuperscript{\rm 3}Data Security Governance Zhejiang Engineering Research Center, Hangzhou Dianzi University, China\\
    \{zehaoliu, mzgao, pjiao\}@hdu.edu.cn
}

\begin{document}
\newtheorem{my_definition}{Definition} 

\maketitle

\begin{abstract}
Multivariate time series anomaly detection has numerous real-world applications and is being extensively studied. Modeling pairwise correlations between variables is crucial. Existing methods employ learnable graph structures and graph neural networks to explicitly model the spatial dependencies between variables. However, these methods are primarily based on prediction or reconstruction tasks, which can only learn similarity relationships between sequence embeddings and lack interpretability in how graph structures affect time series evolution. In this paper, we designed a framework that models spatial dependencies using interpretable causal relationships and detects anomalies through changes in causal patterns. Specifically, we propose a method to dynamically discover Granger causality using gradients in nonlinear deep predictors and employ a simple sparsification strategy to obtain a Granger causality graph, detecting anomalies from a causal perspective. Experiments on real-world datasets demonstrate that the proposed model achieves more accurate anomaly detection compared to baseline methods.
\end{abstract}

%

\section{Introduction}

Time series data are prevalent in the real world. Many industrial systems, such as water supply systems, aerospace, and large server systems, have extensive sensor networks that generate vast amounts of multivariate time series (MTS) data. These systems exhibit complex internal dependencies and nonlinear relationships. With the rapid development in related fields, discovering system anomalies from the large volumes of monitoring data generated by sensors has become an important issue. This is crucial for maintaining the safe operation of systems and reducing economic losses~\cite{TranAD, zhang2024vggm}.

Capturing inter-series relationships has been proven to effectively enhance the performance of MTS anomaly detection~\cite{CorrelationAware, DL4AD}.
Recently, Graph Neural Networks (GNNs) have shown great potential in time series anomaly detection by effectively capturing spatial dependencies between variables~\cite{AD_in_iot}. Considering that most time series datasets do not provide readily available graphs, many existing methods adaptively learn graph structures, and then use reconstruction or prediction errors for anomaly detection~\cite{GNN4TS}.
A key problem is that these models only learn the similarity of sequence embedding vectors without exploring the role of the graph structure in the evolution of time series.

In real-world scenarios, anomalies in time series are often accompanied by changes in dependency structures. For example, consider a simple pipeline in a water supply system. Under normal conditions, when water pressure increases at the inflow end, it leads to a corresponding increase at the outflow end. However, if there is a leak in the pipeline, an increase in inflow pressure may no longer result in the expected pressure increase at the outflow end. 
Therefore, learning interpretable dependency structures with causal relationships can effectively detect anomalies in the evolution patterns of time series. However, there are several key challenges in using causal relationships to detect anomalies:

\begin{itemize}
\item Real systems have complex control logic and numerous nonlinear associations, making it difficult to learn interpretable dependency relationships between variables in a data-driven manner.
\item The dependency relationships between variables are time-varying, and dynamically capturing dependency patterns in the system and identifying anomalies is a key issue.
\end{itemize}

Time series causality is a promising tool for modeling spatial dependencies. 
Granger introduced Granger causality~\cite{Granger1969}.
The core idea of Granger causality is straightforward: if using the series $ x_i $ does not help reduce the prediction error of another series $ x_j $, then $ x_i $ does not Granger-cause $ x_j $~\cite{GrangerSurvey}. The original definition of Granger causality was intended for linear systems, but recent studies~\cite{Granger2019,Causal_NIPS23,CausalTime} have extended Granger causality to nonlinear relationships.

In this paper, we propose a \underline{G}ranger \underline{C}ausality-based multivariate time series \underline{A}nomaly \underline{D}etection method (GCAD). 
To address the first challenge, we combine deep models with Granger causality. Deep networks have powerful modeling capabilities for nonlinear relationships. We leverage this capability to uncover complex causal relationships from black-box deep networks. To tackle the second challenge, we propose using the gradients in nonlinear deep models to dynamically mine causal dependencies. During the training phase, normal causal patterns are embedded into the deep network. When anomalies occur, the nonlinear deep network will yield significantly deviated causal patterns. Our hypothesis is that when an anomaly occurs, the Granger causality patterns between sequences will change significantly.

Based on the definition~\cite{CUTS} of Granger causality, we introduce a channel-separated gradient generator and quantify the Granger causality effect as an integral of deep predictor gradients over the time lag. 
 Our motivation is that the gradients of a network reflect its internal structure to some extent. Utilizing Granger causality discovered from network gradients will help us leverage internal information from black-box models. 
 Discovering Granger causality from the gradient perspective offers many advantages. Deep predictors can automatically learn complex nonlinear relationships between variables. Most importantly, compared to existing Granger causality discovery methods, our approach does not require repeated optimization and parameter adjustment during the testing phase.

Going beyond previous methods, we first train a simple predictor on data without anomalies and then use our proposed gradient-based Granger causality effect quantification method to obtain dynamic spatial dependencies between sequences. To impose sparsity constraints on the causal graph, we use a symmetry-based sparsification method to eliminate bidirectional edges and reduce the impact of sequence similarity on Granger causality effects. It is noteworthy that the output matrix generated by our method contains both spatial dependency information and temporal dependency information. By combining the deviations from these two types of dependencies to compute anomaly scores, our method has achieved excellent results on five real-world benchmark datasets. Our contributions are summarized as follows:

\begin{itemize}

\item 
We propose using the deviation of dynamic Granger causality patterns for time series anomaly detection.
To the best of our knowledge, our proposed method is the first research to combine deep models with Granger causality for time series anomaly detection.

\item We construct Granger causal graphs using deep model gradients, avoiding frequent online optimization during the testing phase, and propose an effective method for causal graph sparsification.

\item Our method achieved state-of-the-art anomaly detection results on most of the five real-world benchmark datasets.

\end{itemize}

\begin{figure*}[t]
\centering
\includegraphics[width=0.82\textwidth]{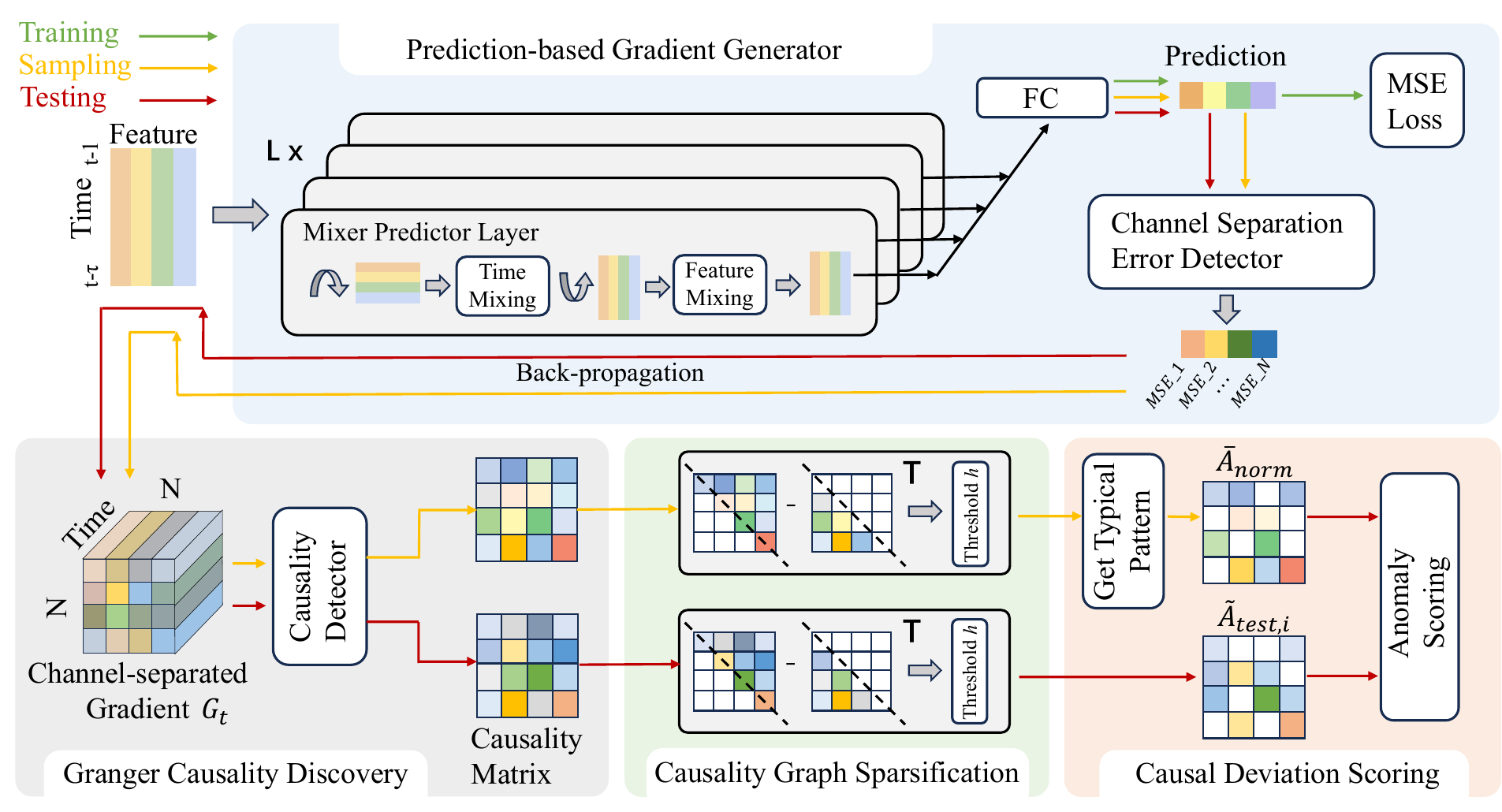}
\caption{Overall Architecture of GCAD. During the training phase, the gradient generator is trained for the prediction task. In the sampling and testing phases, the gradient information from the training samples and test data within the predictor is used to perform Granger causality discovery, and the causal graphs are obtained through sparsification. Anomaly scores are calculated by measuring the deviation of the causal graphs from the normal pattern.}
\label{fig1:model}
\end{figure*}

\section{Related Work}

\subsection{Multivariate Time Series Anomaly Detection}

Multivariate time series anomaly detection is widely applied in the real world. As a classic task in the field of time series analysis, it has received extensive research attention. 
Early research primarily focused on statistical methods, such as ARIMA~\cite{ArimaMethod}, \cite{hotsax} and~\cite{distance2018}. Recently, many deep learning-based methods have been proposed, which effectively capture nonlinear information and are not constrained by the stationarity assumption. These include methods based on Convolutional Neural Networks (CNN)~\cite{CNNMethod2018,CNNMethod_2,CNNLSTM} and Long Short-Term Memory(LSTM) networks~\cite{SMAP}, among others.

With the development of autoencoders and their variants, many studies have explored using Variational Autoencoders (VAE)~\cite{VAEMethod,smd} or Generative Adversarial Networks (GAN)~\cite{BeatGAN,MAD_GAN} for anomaly detection. However, these methods do not explicitly learn the spatial dependencies between sequences, and as pointed out in~\cite{CorrelationAware}, they fail to fully exploit the dependencies between variable pairs.

Recently, graph neural networks (GNN) have been increasingly applied to the problem of sequence anomaly detection. By modeling sequences as nodes and the correlations between sequences as edges, using a graph data structure and GNNs naturally represents the spatial relationships between variables. A key challenge in applying GNNs to sequence anomaly detection is that the required knowledge of graph structure usually does not inherently exist in time series anomaly detection data~\cite{structure_available}. To address this challenge, 
many studies have explored adaptive graph learning modules, with GDN~\cite{GDN} being a pioneering work in this area~\cite{GNN4TS}. GDN calculates the graph structure based on the cosine distance of sequence embeddings. Most graph-based methods~\cite{FuSAGNet} use GNNs for prediction or reconstruction, but they do not directly apply spatial dependency patterns to anomaly detection.

\subsection{Inter-sequence Correlations Modeling}

Explicitly modeling spatial dependencies between sequences helps in discovering more complex anomalies. Most GNN-based methods construct static graph structures using randomly initialized embedding vectors to model spatial dependencies. Methods like VGCRN~\cite{VGCRN} and FuSAGNet~\cite{FuSAGNet} compute dot products between embeddings to generate a similarity matrix used as a graph structure. However, optimization methods based on downstream tasks may not yield stable and meaningful graph structures. Another class of methods leverages self-attention mechanisms for modeling. The reconstruction module of Grelen~\cite{Grelen} learns to dynamically construct graph structures that adapt to each time point based on the input time series data. However, randomly initialized attention networks may result in learned spatial relationships that lack practical significance~\cite{Space_modeling_RW}.

Some limited studies~\cite{Granger_AD_2012} have explored using Granger causality to model spatial relationships between sequences, but they are based on simple linear statistical models and cannot capture complex nonlinear dependencies. 
Other methods~\cite{Granger2021} discover Granger causality from sparse penalized network weights. However, this approach is not suitable for anomaly detection tasks. Because data is input in a streaming fashion, and these methods have to continuously optimize parameters during testing to adapt to dynamic causal relationships, resulting in substantial computational costs.

\section{Methodology}
\subsection{Problem Statement}

In this paper, we investigate the task of anomaly detection in multivariate time series. Our multivariate time series data are collected from various sensors in a real system, observed at equal intervals over a period of time. The observed time series can be represented as a set of time points: $ \{x_1, x_2, ..., x_T\} $, where $ x_t \in \mathbb{R}^N $ denotes the observations from $ N $ sensors at time $ t $. In the anomaly detection task, the input to the model is a sliding window $\{x_{t-\tau},...,x_{t-1},x_{t}\}$.
The model outputs a Boolean value for each sliding window to determine whether there is an anomaly within that window.

\subsection{Overview}

Our GCAD framework aims to extract Granger causality relationships among multivariate time series and then identify anomalies from the causal patterns on the test set. It mainly consists of the following four parts:

\begin{enumerate}
    \item \textbf{Prediction-based Gradient Generator:} Utilizes predictive methods to guide training and provides channel-separated gradients during the causality discovery phase.
    \item \textbf{Granger Causality Discovery:} Dynamically infers Granger causal relationships from the gradients produced by the gradient generator.
    \item \textbf{Causality Graph Sparsification:} Applies sparsity constraints to the discovered causal relationships to obtain a causality graph matrix.
    \item \textbf{Causal Deviation Scoring:} Calculates the causal pattern deviation score and integrates temporal information to detect anomalies.
\end{enumerate}

Figure~\ref{fig1:model} provides an overview of our framework.

\subsection{Prediction-based Gradient Generator}

The widely studied Mixer predictor~\cite{tsmixerGoogle} is used as the gradient generator in our GCAD framework. The predictor consists of $ L $ stacked Mixer Predictor Layers, each containing interleaved temporal mixing and feature mixing MLPs. The temporal mixing MLPs are shared across all $ N $ features, while the feature mixing MLPs are shared across all time steps. The output of each layer is fed through skip connections into a fully connected layer to produce the predictive output.

The input to the predictor is the sliding window $ X_{t-1} = \{x_{t- \tau}, x_{t- \tau +1}, ..., x_{t-1}\} $, where $ \tau $ is the maximum time lag considered for Granger causality, and $X_t \in \mathbb{R}^{N \times \tau}$. The predictor outputs the prediction for time $t$: $\widehat{y_t} = f(X_{t-1})$. Where $f$ is the prediction function fitted by the predictor, and $ \widehat{y_t} \in \mathbb{R}^N $. During the training phase, the MSE (Mean Squared Error) loss is used to guide the optimization of predictor parameters: $ L_{train} = MSE(\widehat{y_t}, y_t) $, where $ y_t $ is the ground truth.

During the testing phase, in order to explore the causal relationships between variables, the gradient generator needs to compute pairwise predictor gradients between variables. Therefore, we propose a channel-separated Error detector to generate channel loss $ L_t \in \mathbb{R}^N $:
\begin{equation}
\begin{aligned}
L_{t,j} = (\widehat{y}_{t,j} - y_{t,j})^2 \textnormal{,}
\end{aligned}
\label{channelErr}
\end{equation}where $ L_{t,j} $ represents the prediction error of sequence $ j $ in the sliding window $ X_t $. Next, the gradient generator performs backpropagation on each prediction error through the prediction network, obtaining the gradient $ G_{t,j} \in \mathbb{R}^{N \times \tau} $ on $ X_t $. We compile the complete gradient tensor $ G_t \in \mathbb{R}^{N \times N \times \tau} $ by stacking all the gradients together.

\subsection{Granger Causality Discovery}

Nonlinear Granger causality is defined from the perspective of the impact of variables on each other's predictive effects. Our framework reconsiders Granger causality from the perspective of gradients in deep networks. In this paper, we adopt the widely used definition~\cite{CUTS} of nonlinear Granger causality:

\begin{my_definition}

Time series $ i $ Granger-causes $ j $ if and only if there exists $ x'_{t-\tau : t-1 , i} \neq x_{t-\tau : t-1 , i} $,

\begin{equation}
\begin{aligned}
    &f_j (x_{t-\tau : t-1 , 1}, ... , x'_{t-\tau : t-1 , i}, ... ,x_{t-\tau : t-1 , N}) \neq 
    \\ &f_j (x_{t-\tau : t-1 , 1}, ... , x_{t-\tau : t-1 , i}, ... ,x_{t-\tau : t-1 , N}) \textnormal{.}
\end{aligned}
\label{GrangerDef}
\end{equation}

i.e., the past data points of time series $ i $ influence the prediction of $ x_{t,j}. $
    
\label{Granger_Definition}
\end{my_definition}

Considering this definition from a differential perspective, let $ t' \in (t-\tau:t-1) $, and let $ x^*_{t',i} $ be a perturbation of $ x_{t',i} \in X_t $:
\begin{equation}
\begin{aligned}
x^*_{t',i} = x_{t',i} + \Delta \textnormal{,}
\end{aligned}
\label{DeltaX}
\end{equation}where $ \Delta $ is the perturbation. Based on the predictor described in the previous subsection, the following equations can be obtained: $\\ \widehat{y}_{t,j} = f_j (x_{t-\tau : t-1 , 1}, ... , x_{t-\tau : t-1 , i}, ... ,x_{t-\tau : t-1 , N}) $, and $\widehat{y^*}_{t,j} = 
f_j (x_{t-\tau : t-1 , 1} ... , \{ x_{t-\tau, i}, ...,x*_{t', i}, ...,x_{t-1,i} \}, ... , \\x_{t-\tau : t-1 , N}) $. Where $ f_j $ is the function of $\widehat{y}_{t,j}$ with respect to the input in the prediction network. The change in prediction error caused by the perturbation can be transformed into the form of partial derivatives:
\begin{equation}
\begin{aligned}
\lim_{\Delta \to 0} \lvert L_{t,j} - L^*_{t,j} \lvert = \left\vert \frac{\partial L_{t,j}}{\partial x_{t',i}} \right\vert \cdot \left\vert \Delta \right\vert\textnormal{,}
\end{aligned}
\label{DeviationPartial}
\end{equation}where $ L^*_{t,j} = (\widehat{y^*}_{t,j} - y_{t,j})^2 $. Granger causality considers the mutual influence of sequences on each other's predicted values within the maximum time lag. Therefore, we define the quantification of Granger causality as the integral of the absolute values of channel-separated gradients over the time lag:
\begin{equation}
\begin{aligned}
a_{i,j} = \int_{t- \tau}^{t-1} \left\vert \frac{\partial L_{t,j}}{\partial x_{\phi,i}} \right\vert P(x_{\phi,i}) d x_{\phi,i} \textnormal{,}
\end{aligned}
\label{aij}
\end{equation}where $\phi$ is the time index from $t-\tau$ to $t-1$.The term $ a_{i,j} $ represents the degree to which sequence $ i $ Granger causes sequence $ j $, parameterized by a distribution of interest $ P $. The causality matrix is defined as $ A = \{ a_{i,j} \}^N_{i,j=1} $.

Since predictors composed of deep networks are capable of backpropagation, the prediction function $ f $ is inherently continuous and differentiable. For simplicity, the interest distribution $ P $ can be a uniform distribution.
According to Equation~\ref{DeviationPartial}, when $ a_{i,j} \neq 0 $, it can be concluded that the two predicted values are not equal, i.e., $\widehat{y^*}\_{t,j} \neq \widehat{y}\_{t,j}$.
Since the inputs corresponding to these two predicted values are $x^*_{t',i}$ and $x_{t',i}$, respectively, the differing inputs of sequence $i$ result in differing outputs of sequence $j$. This is consistent with Definition~\ref{Granger_Definition}, from which it can be inferred that sequence $i$ Granger-causes sequence $j$.


\subsection{Causality Graph Sparsification}

Unlike similarity or correlation between sequences, causality must be unidirectional. The ideal Granger causality graph is a directed acyclic graph (DAG). However, in nonlinear Granger causality discovery, it is difficult to strictly guarantee the acyclic nature of the causal graph. Existing methods discover nonlinear causality through network weights constrained by sparsity. 
To adapt to the anomaly detection task, instead of directly constraining the weights, we employ sparsification to address the dynamic Granger causality identified from the gradients of the deep network.

The causal relationships discovered from the gradients of deep networks may include undirected edges, which to some extent represent the similarity between sequences. Our intuition is that this similarity should be equal in both directions. In fact, the widely used relationship matrices obtained from cosine similarity are symmetric matrices, thus assuming symmetry in both directions has convincing empirical evidence. Based on this intuition, we propose a simple and easy-to-use method for sparsifying causality graphs:
\begin{equation}
\begin{aligned}
\widetilde{A}_{i,j} &= max(0, A{i,j} - A^T_{i,j}), i \neq j \textnormal{,}\\
\widetilde{A}_{i,i} &= A_{i,i} \textnormal{.}
\end{aligned}
\label{sparsification}
\end{equation}

In which $ \widetilde{A} $ is the sparsified Granger causality graph matrix. The causality matrix subtracted from its transpose eliminates bidirectional symmetric similarities while preserving unidirectional Granger causality. Further, we set a sparsity threshold $ h $, setting causality effect values below this threshold in the causality graph matrix to zero. This is because insignificant causal relationships may be caused by noise and are not helpful for anomaly detection.

\subsection{Causal Deviation Scoring}

By sliding a window over the test set and using the proposed framework to construct Granger causality graphs, we obtain a sequence of causality graphs. We aim to detect anomalies that deviate from the normal causal patterns within these graphs. A straightforward idea is to construct a causality graph that represents the normal causal pattern and compare each causality graph from the test set against it. To leverage the causal pattern information from the normal data, after model training, we sample the training set windows using a Bernoulli distribution and calculate the Granger causality graphs for these samples. Then, we use the mean matrix of the graph matrix sequence to represent the typical normal causal pattern:
\begin{equation}
\begin{aligned}
W'_{train} &= W_{train} \odot B \textnormal{,}\\
B &= \{ b_1. ... , b_{n_{train}} \} \textnormal{,}\\
b_i &\thicksim Bernoulli(p) \textnormal{,}\\
A_{norm, i} &= \{ g(W'_{train, i}) \}^n_{i=1} \textnormal{,}\\
\end{aligned}
\label{normA}
\end{equation}where $ n_{train} $ is the total number of sliding windows in the training set, $ p $ is the parameter of the Bernoulli distribution, $ n $ is the number of sampled windows, and $ g $ is the function used to compute the causality graph. The resulting typical causality pattern matrix $ \overline{A}_{norm} \in \mathbb{R}^{N \times N} $:
\begin{equation}
\begin{aligned}
\overline{A}_{norm} &= \frac{1}{n} \sum\limits_{i=1}^{n} \widetilde{A}_{norm, i} \textnormal{,}
\end{aligned}
\label{averageA_norm}
\end{equation}where $ \widetilde{A}_{norm} $ is the sparsified causal matrix, and $ n $ is the total number of samples obtained through sampling.

Calculate the Granger causality graph for each window in the test set $ W_{test} $:
\begin{equation}
\begin{aligned}
\widetilde{A}_{test} = \{ g(W_{test,i}) \}^n_{i=1} \textnormal{.}
\end{aligned}
\label{testA}
\end{equation}

Define the causal deviation score for each test sample as:
\begin{equation}
\begin{aligned}
Sc_i = \sum \frac{\left\vert \widetilde{A}_{test,i} - \overline{A}_{norm} \right\vert}{\overline{A}_{norm} + \varepsilon} \textnormal{,}
\end{aligned}
\label{causalScore}
\end{equation}where $ \varepsilon $ is a very small value, and $ Sc $ represents the sequence of causal deviation scores. 
According to Equation~\ref{aij}, the main diagonal of the causality graph matrix represents the extent to which variables influence their own predicted values within the maximum time lag. To some extent, the values on the main diagonal reflect time-dependent patterns, although they do not explicitly provide the specifics of time dependence. We define the time pattern deviation as follows:
\begin{equation}
\begin{aligned}
St_i = \sum \frac{\left\vert diag \left( \widetilde{A}_{test,i} - \overline{A}_{norm} \right) \right\vert}{diag \left(\overline{A}_{norm} \right) + \varepsilon } \textnormal{.}
\end{aligned}
\label{timeScore}
\end{equation}

$ St $ is the sequence of time pattern deviations, and $ diag(\cdot) $ refers to forming a diagonal matrix from the main diagonal elements. The final anomaly score is composed of a mixture of causal pattern deviations and time pattern deviations:
\begin{equation}
\begin{aligned}
S = Sc + \beta St \textnormal{,}
\end{aligned}
\label{AnomalyScore}
\end{equation}where $\beta$ is a hyperparameter that balances the causal pattern deviation and the time pattern deviation.Note that the causal pattern deviation score inherently includes the main diagonal information, so even when $ \beta $ is set to 0, it still contains time pattern information.

\section{Experimental Results}

To evaluate our proposed GCAD framework, we conducted experiments on five widely-used real-world benchmark datasets and compared it with six popular baseline methods.

\subsection{Experimental Setup}
\noindent\textbf{Datasets.}
Experiments are conducted on five real-world datasets, including SWaT~\cite{swat}, SMD~\cite{smd}, MSL, SMAP~\cite{SMAP}, and PSM~\cite{PSM}. The statistical information of these datasets is presented in Table~\ref{Datasets}.

\noindent\textbf{Baseline Methods.}
The baseline methods include DAGMM~\cite{DAGMM}, USAD~\cite{USAD}, GDN~\cite{GDN}, AnomalyTransformer(AT)~\cite{AT}, GANF~\cite{GANF} and MEMTO~\cite{MEMTO}.

\noindent\textbf{Implementation Details.}
Each dataset consists of two parts: unlabeled normal operation data and labeled data containing some anomalies. We use 80\% of the normal data for training, and the remaining 20\% is used for the validation set. Testing is conducted on the data containing anomalies. Since most baseline methods do not provide a way to set predetermined thresholds, we evaluate using two threshold-independent metrics: the Area Under the Curve (AUC) of the Receiver Operating Characteristic (ROC) and the Precision-Recall Curve (PRC). All experiments were conducted 10 times and the average results were reported.

\begin{table}[t]
  \centering
    \begin{tabular}{l|c|c|c|c}
    \toprule
    Dataset & channels & train & test & anomalies \\
    \midrule

    SWaT & 51 & 47,520 & 44,991 & 12.20\% \\
    SMD & 38 & 28,479 & 28,479 & 9.46\% \\
    MSL & 55 & 3,682 & 2,856 & 0.74\% \\
    SMAP & 25 & 2,876 & 8,579 & 2.14\% \\
    PSM & 25 & 132,481 & 87,841 & 27.76\% \\

    \bottomrule
    \end{tabular}
  \caption{Statistics of the Datasets}
  \label{Datasets}
\end{table}

\begin{table*}[htbp]
  \centering
  \resizebox{.98\textwidth}{!}{
    \begin{tabular}{l|cc|cc|cc|cc|cc}
    \toprule
    Dataset & \multicolumn{2}{c}{SWaT} & \multicolumn{2}{c}{SMD} & \multicolumn{2}{c}{MSL} & \multicolumn{2}{c}{SMAP} & \multicolumn{2}{c}{PSM}  \\
    Metric & ROC   & PRC    & ROC  & PRC    & ROC   & PRC    & ROC   & PRC    & ROC   & PRC    \\
    \midrule

    DAGMM & 0.7882 & 0.4955 & 0.7516 & 0.3988 & 0.6209 & 0.0163 & 0.5989 & 0.0380 & 0.6556 & 0.3860 \\
    USAD & 0.8318 & \underline{0.7173} & \underline{0.9274} & 0.5316 & 0.5601 & 0.0108 & 0.5314 & 0.0279 & 0.6584 & 0.4924 \\
    GDN & \underline{0.8493} & 0.6076 & 0.9006 & \underline{0.5655} & 0.3846 & 0.0055 & 0.5115 & 0.0338 & \underline{0.7284} & \underline{0.4964} \\
    AT & 0.5117 & 0.1851 & 0.1765 & 0.0576 & 0.5391 & 0.0141 & 0.3467 & 0.0188 & 0.5058 & 0.2902 \\
    GANF & 0.8112 & 0.3557 & 0.6384 & 0.0017 & 0.6402 & 0.0291 & \underline{0.6926} & \textbf{0.5259} & 0.6335 & 0.4090 \\
    MEMTO & 0.7799 & 0.6067 & 0.5042 & 0.1187 & \underline{0.7271} & \underline{0.0654} & 0.4791 & 0.0176 & 0.5026 & 0.2913 \\
    
    \midrule
    GCAD & \textbf{0.8690} & \textbf{0.7758} & \textbf{0.9533} & \textbf{0.7502} & \textbf{0.7658} & \textbf{0.3679} & \textbf{0.7273} & \underline{0.4555} & \textbf{0.7618} & \textbf{0.6136} \\
    \bottomrule
    \end{tabular} } 
  \caption{Anomaly detection accuracy in terms of AUROC and AUPRC, on five benchmark datasets with ground-truth labelled anomalies}
  \label{overall}
\end{table*}

\subsection{Anomaly Detection Performance}

The results of our framework and the six baseline methods are summarized in Table~\ref{overall}.

From the results, it can be seen that GCAD achieves state-of-the-art (SOTA) performance in most cases. DAGMM and USAD are classical anomaly detection frameworks that do not explicitly model the spatial relationships between sequences, which limits their anomaly detection performance. GDN achieves the second-best performance across four experimental metrics; however, it uses an adaptive graph structure learning strategy to learn a fixed graph structure, which is not suitable for dynamically changing systems. GANF achieves the best result on the PRC metric for the SMAP dataset. This is because the SMAP dataset has severe distribution shifts, and GANF, being a density-based method, can better handle this situation by learning the evolution of graph structures and identifying distribution shifts. It is worth noting that the MSL and SMAP datasets are relatively small and have severe class imbalance, which may lead to very low PRC values for most methods. MEMTO, by incrementally training individual items in a gated memory module, mitigates the overgeneralization problem of reconstruction-based models. Therefore, it achieves the second-best result on the MSL dataset, where the proportion of anomalous samples is extremely small.

\subsection{Ablation Study}

We investigated the effect of each component in the proposed framework. Table~\ref{ablation_table} shows the results of the ablation study on two datasets. 
"-Spars" indicates the removal of the Causality Graph Sparsification part from GCAD. "-GC" means not using Granger causality for anomaly detection. "-TC" means disregarding the temporal correlations within each sequence by excluding the time pattern deviation from the anomaly score.

The results show that causal graph sparsification brings certain performance improvements. This is because sparsification helps mitigate the impact of sequence similarity on causal relationship discovery. 
Furthermore, data from large-scale systems typically consist of numerous channels. Sparsification helps reduce noise in the identified causal relationships.

The introduction of Granger causality and temporal correlations has led to significant performance improvements. This demonstrates that our model effectively captures the spatial and temporal associations in time series data. Overall, Granger causality more noticeably enhances our model's performance, indicating that using causal relationships for anomaly detection in real-world datasets is important and effective.

\subsection{Effect of Parameters}

\begin{table}[t]
  \centering
    \begin{tabular}{l|cc|cc}
    \toprule
    Dataset & \multicolumn{2}{c}{SWaT} & \multicolumn{2}{c}{SMD} \\
    Metric & ROC & PRC & ROC & PRC \\
    \midrule
    GCAD & \textbf{0.8690} & \textbf{0.7758} & \textbf{0.9533} & \textbf{0.7502} \\
    -- Spars & 0.8559 & 0.7676 & 0.9531 & 0.6915 \\
    -- GC & 0.8478 & 0.7513 & 0.9405 & 0.6424 \\
    -- TC & 0.8400 & 0.7558 & 0.9498 & 0.6748 \\
    \bottomrule
    \end{tabular}
  \caption{Ablation study. The anomaly detection results with and without components in GCAD}
  \label{ablation_table}
\end{table}

\begin{figure}[b]
\centering
\begin{subfigure}[t]{0.23\textwidth}
\centering
\includegraphics[width=\textwidth]{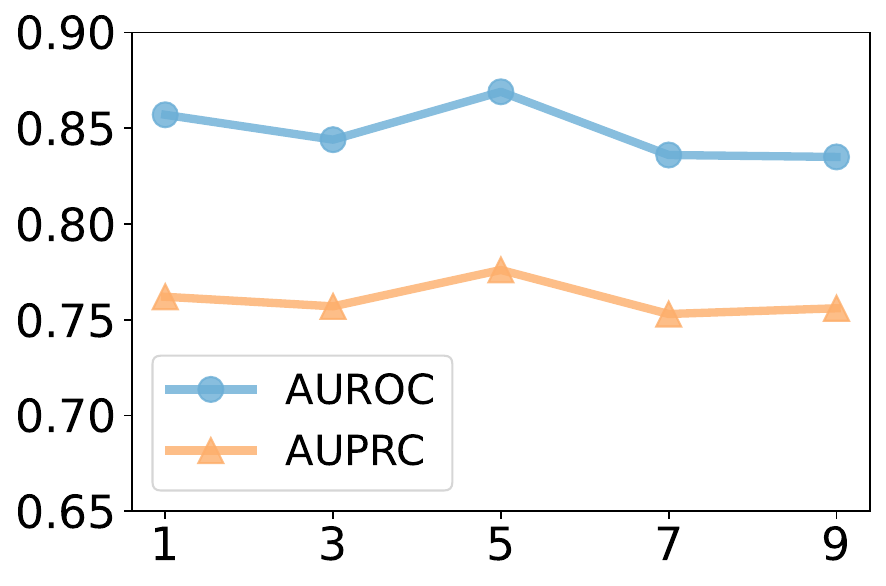} 
\caption{Max Time Lag}
\end{subfigure}
\begin{subfigure}[t]{0.23\textwidth}
\centering
\includegraphics[width=\textwidth]{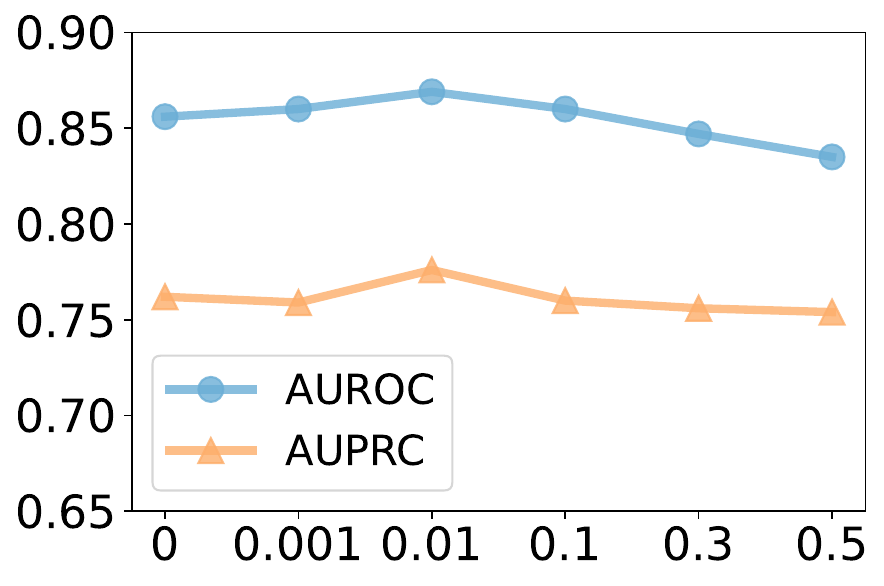}
\caption{Sparsification Threshold}
\end{subfigure}
\caption{Effect of parameters. AUROC and AUPRC as functions of (a) maximum time lag $ \tau $ in Granger causality and (b) sparsification threshold $ h $ of the causal graph}
\label{parameters_fig}
\end{figure}

We further investigated the impact of key hyperparameters on the anomaly detection performance of GCAD. All experiments were conducted using the SWaT dataset, and the results are shown in Figure~\ref{parameters_fig}.

The maximum time lag $ t $ determines how many lagged effects of Granger causality the model can discover. When $ t $ is set to 1, the model only discovers Granger causality in adjacent time steps, and the shorter sliding window makes the model more sensitive to anomalies within the window. As $ t $ increases, the model can discover Granger causality with higher lags, which helps extract more complex high-order spatial patterns. However, this also reduces the model's sensitivity to short-term anomalies, because GCAD focuses on the causal patterns in the input distribution over the entire window (Equation~\ref{aij}).
Therefore, the maximum time lag parameter controls the balance between the model's sensitivity to anomalies and its ability to discover complex anomalies. A moderate $ t $ value can achieve better performance.

The sparsification parameter $ h $ determines the threshold below which causal relationships are considered as noise. When $ h $ is small, there are many non-zero but insignificant causal relationships in the causality matrix. 
When $ h $  is too large, GCAD focuses only on the most significant causal relationships, ignoring the broader causal relationships that exist in the system, thus failing to fully utilize the anomaly information contained in the causal patterns. Experiments show that a moderate $ h $ can yield better results.

\subsection{Analysis of Anomaly Detection Examples}

\begin{figure}[t]
\centering
\includegraphics[width=0.45\textwidth]{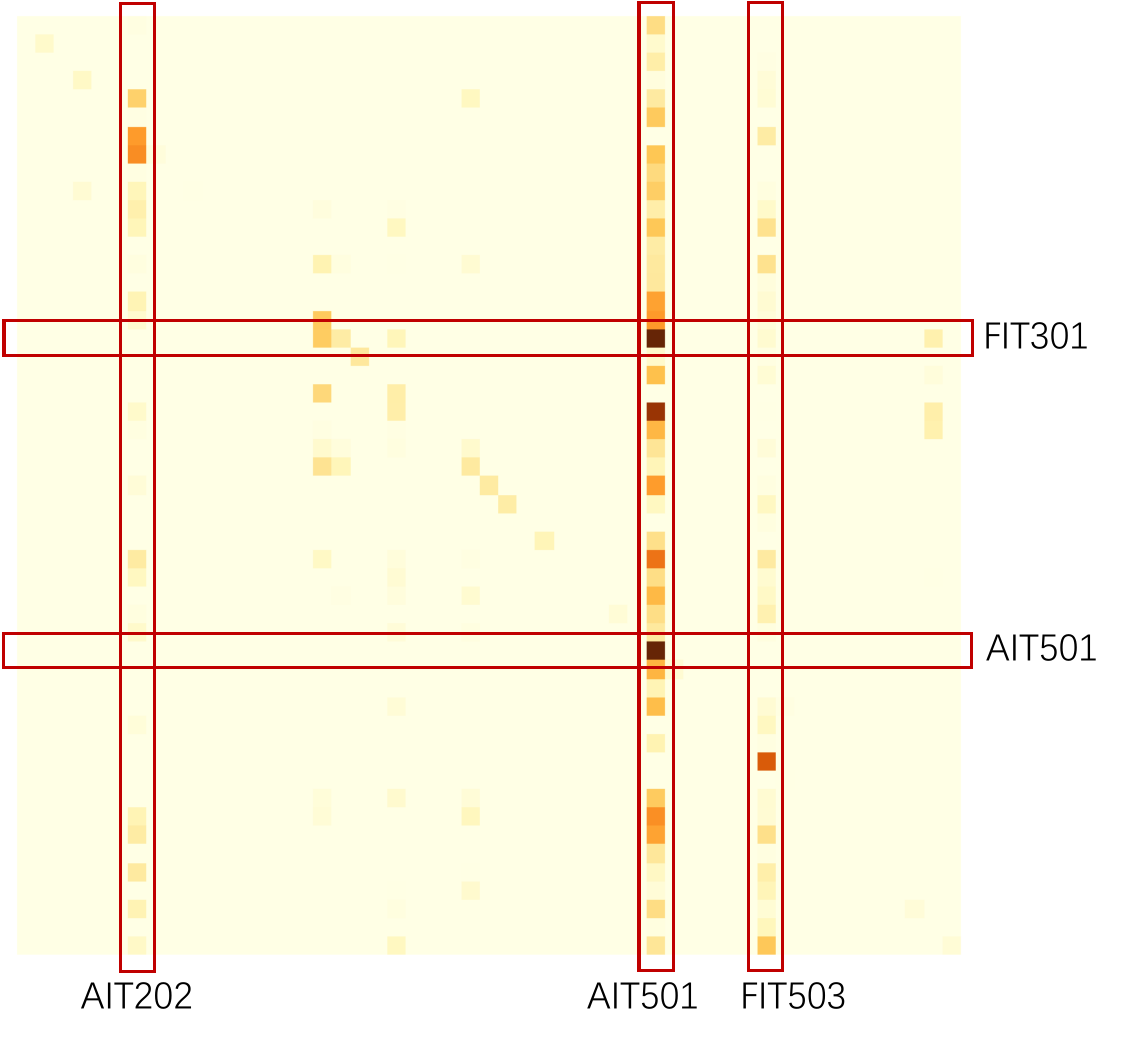}
\caption{Causal Pattern Deviation Matrix in an Example Anomaly on the SWaT Dataset}
\label{swat_attack_causal}
\end{figure}

\begin{figure}[h]
\centering
\includegraphics[width=0.48\textwidth]{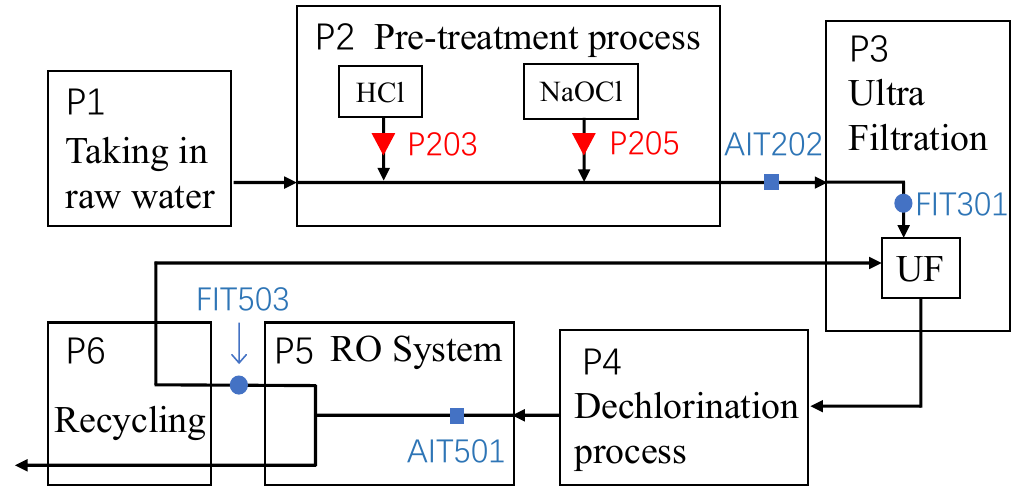}
\caption{Physical Structure of the SWaT Testbed, Attack Points of Anomalous Events (marked in red), and Main Affected Points (marked in blue)}
\label{swat_structure}
\end{figure}

\begin{figure}[t]
\centering
\begin{subfigure}[t]{0.4\textwidth}
\centering
\includegraphics[width=\textwidth,height=0.06\textheight]{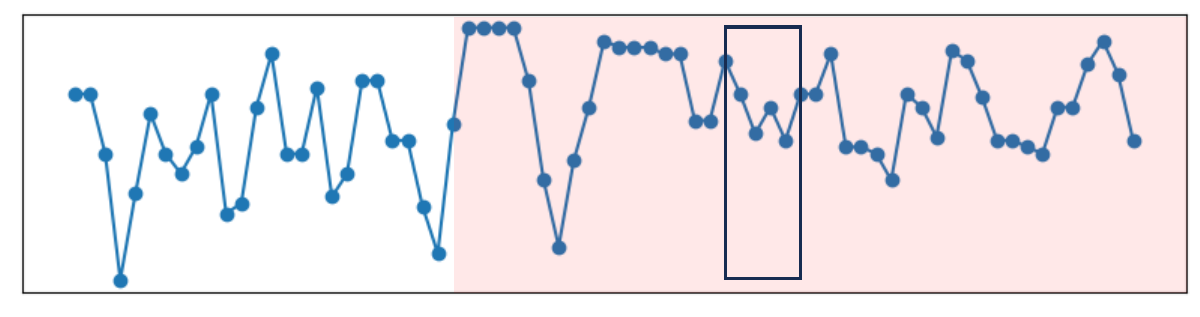}
\caption{AIT202}
\end{subfigure}
\begin{subfigure}[t]{0.4\textwidth}
\centering
\includegraphics[width=\textwidth,height=0.06\textheight]{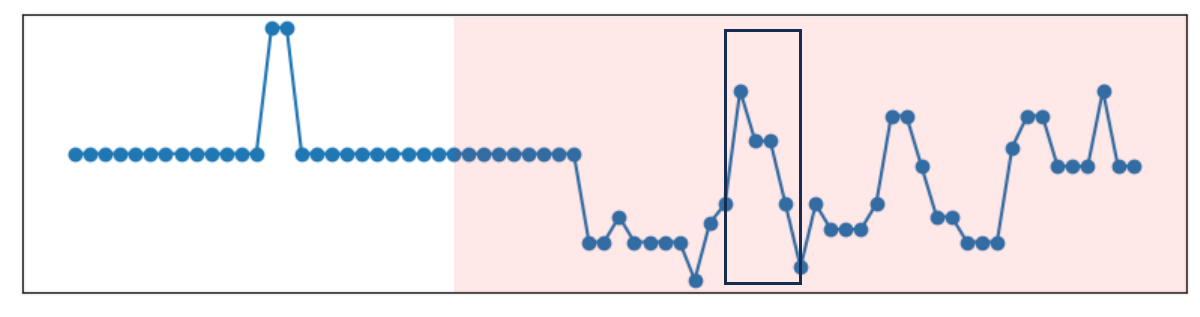}
\caption{AIT501}
\end{subfigure}
\begin{subfigure}[t]{0.4\textwidth}
\centering
\includegraphics[width=\textwidth,height=0.06\textheight]{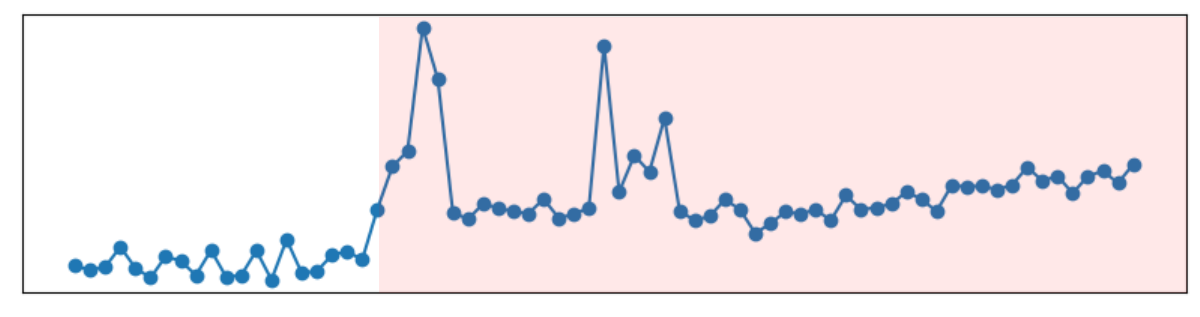}
\caption{Anomaly Score}
\end{subfigure}
\caption{Two Affected Sensors and the Changes in the Total Anomaly Score in the Example Anomalous Instances}
\label{anomaly_score_fig}
\end{figure}

To demonstrate how causal patterns reveal anomalies, we conducted a case study on anomaly event data from a real system. The experiments were performed on the SWaT dataset, as the original authors~\cite{swat_structure} provided the physical construction and sensor descriptions of this real system.

To facilitate observation, we visualized the causal pattern deviations provided by GCAD during the occurrence of anomaly events, as shown in Figure~\ref{swat_attack_causal}. In this figure, the deviation matrix $ D $ represents the absolute error between the causality matrix during the anomaly event and the typical causality pattern matrix, that is, $ D = \left\vert \widetilde{A}_{test,i} - \overline{A}_{norm} \right\vert$. Darker colors indicate larger deviations. We have marked the main affected points in this attack in Figure~\ref{swat_attack_causal}, and Figure~\ref{swat_structure} illustrates the physical relationships between these points.

Specifically, this anomaly attack targeted P203 and P205, which are two pumps in Process 2 responsible for the addition of HCl and NaOCl, respectively. The attackers maliciously shut down these two pumps, causing changes in water quality. AIT202, located downstream of the attack points, is a sensor that measures the concentration of HCl in the water. The most significantly affected sensor is AIT501, located in Process 5 downstream, which is another sensor measuring HCl concentration. When the HCl injection pumps were attacked, the two sensors related to HCl exhibited significant deviations in their causal patterns, indicating that GCAD can effectively detect anomalous changes in causal patterns. Notably, the main diagonal element value corresponding to AIT501 is relatively large, which means AIT501 has a significant temporal pattern deviation for itself, highlighting the importance of incorporating temporal pattern deviations into causal pattern deviations. The other two points with notable causal deviations are FIT301 and FIT503, which are flow meters. 
FIT301 measures the input flow to the UF module, and FIT503 measures the RO Reject flow. 
The deviations at these two points arise from changes in the relationships between certain flows and other sensors in the system under attack.

Figure~\ref{anomaly_score_fig} shows the changes in the two sensors measuring HCl concentration and the variations in anomaly scores during the attack. The red areas indicate the presence of anomaly events, and the causality matrix is derived from the marked sliding window. It can be observed that anomalies in a single sequence are difficult to identify manually. This is because real-world systems like SWaT have complex system dynamics and control logic, where anomalous events may not cause significant fluctuations in the time series. However, GCAD captures the causal relationships between sequences and provides significant changes in anomaly scores by detecting deviations in Granger causal patterns.

\section{Conclusion}
In this work, we propose a Granger Causality-based multivariate time series Anomaly Detection method (GCAD). This framework models the spatial dependencies between sequences using Granger causality, constructs dynamic causal relationships based on the gradients of a deep predictor, and improves the causal graph structure using our sparsification strategy. Experiments on five real-world sensor datasets demonstrate the superiority of GCAD in anomaly detection accuracy compared to other baseline methods. This study explores the integration of Granger causality with deep networks, offering a new perspective for time series anomaly detection. 
However, GCAD also has some limitations, as it can only capture binary causal pairs between variables and cannot model multivariable interactions. Future work may consider multivariable causal relationships to uncover more complex anomalies.

\section{Acknowledgments}
This work was supported in part by the Zhejiang Provincial Natural Science Foundation of China under Grant LDT23F01012F01,  in part by the Zhejiang Provincial Natural Science Foundation under Grant MS25F030032 and in part by the National Natural Science Foundation of China under Grants 62372146 and 62003120.


\bibliography{aaai25}

\end{document}